# A New Unified Method for Detecting Text from Marathon Runners and Sports Players in Video

[1]Sauradip Nag, [2]Palaiahnakote Shivakumara, [3]Umapada Pal, [4]Tong Lu and [5]Michael Blumenstein

[1]Kalyani Government Engineering College, Kalyani, Kolkata, India

[2]Faculty of Computer Science and Information Technology, University of Malaya, Kuala Lumpur, Malaysia.

[3]Computer Vision and Pattern Recognition Unit, Indian Statistical Institute, Kolkata, India.

[4]National Key Lab for Novel Software Technology, Nanjing University, Nanjing, China.

[5]University of Technology Sydney (UTS), Sydney, Australia.

Email: sauradipnag95@gmail.com, shiva@um.edu.my, umapada@isical.ac.in, lutong@nju.edu.cn, michael.blumenstein@uts.edu.au.

**Abstract**

Detecting text located on the torsos of marathon runners and sports players in video is a challenging issue due to poor quality and adverse effects caused by flexible/colorful clothing, and different structures of human bodies or actions. This paper presents a new unified method for tackling the above challenges. The proposed method fuses gradient magnitude and direction coherence of text pixels in a new way for detecting candidate regions. Candidate regions are used for determining the number of temporal frame clusters obtained by K-means clustering on frame differences. This process in turn detects key frames. The proposed method explores Bayesian probability for skin portions using color values at both pixel and component levels of temporal frames, which provides fused images with skin components. Based on skin information, the proposed method then detects faces and torsos by finding structural and spatial coherences between them. We further propose adaptive pixels linking a deep learning model for text detection from torso regions. The proposed method is tested on our own dataset collected from marathon/sports video and three standard datasets, namely, RBNR, MMM and R-ID of marathon images, to evaluate the performance. In addition, the proposed method is also tested on the standard natural scene datasets, namely, CTW1500 and MS-COCO text datasets, to show the objectiveness of the proposed method. A comparative study with the state-of-the-art methods on bib number/text detection of different datasets shows that the proposed method outperforms the existing methods.

**Keywords**: Video text analysis, Gradient direction, Bayesian classifier, Face detection, Torso detection, Deep learning, Text detection.

## 1. Introduction

As the internet expands and the use of social media increases, the rise of easily accessible multimedia data such as video, images and texts have created an unprecedented information overflow. In contrast to text and image media, retrieving video from large datasets, especially from marathons and sports events, is currently in demand [1, 2]. This is because recognizing and retrieving short videos rather than full videos, such as highlights in the case of sports, and finding suspicious behaviors of persons in marathon videos [3], are drawing special attention. In addition, sports and marathons have become an integral part of human life in terms of entertainment and awareness, respectively. In these multimedia applications, those methods that use text as one mode and video/audio as another mode for deriving semantic relationships between different modalities can retrieve the desired events from video. Since semantics of text in images is closer to content of video, those methods that use text information as a kind of high-level feature and further combine it with other features to understand the meaning of video or annotate images, will be useful. In addition, to track a player in sports video to understand his/her behaviors and activities, it is common to explore text information of the player along with other features. Therefore, one can argue that text detection is an integral part of multimedia processing for the present and future applications. However, it is noted that detecting the text of a marathon runner and sports player is challenging compared to text detection in natural scenes. This is because text on the body of a marathon runner or sports player suffers largely from arbitrary-shaped characters due to cloth folding or partial occlusion with human actions, and the text itself may overlap with the clothing color. As a result, one can infer that detecting text from marathon runners and sports players in video is complex and interesting in the text detection field [4, 5].

In recent studies, many powerful text detection methods have been developed through exploring deep learning techniques in different ways to address complex issues such as arbitrary orientation of text, multiple scripts, fonts or font sizes, low contrast, and complex backgrounds. For instance, the methods for natural scenes [6-20], video images [21-23] and text tracking with the help of temporal frames [24-27] have been developed in the literature. However, it is noted that these methods may not be adequate to handle the above-mentioned challenges of marathon and sports video. This is because color of skin and some parts of human bodies may overlap with those of text information. This results in more false positives or the loss of text. It is evident because in the case of marathon and sports video, since the human body is considered as prominent information, one cannot expect the text of marathon runners or sports players to provide high quality information. In addition, since it is not focused information in video, one can expect text affected by several challenges. In the case of natural scene images, since text is considered as prominent information, one can expect that text can preserve good quality, while other information can have low quality. Therefore, the methods developed for natural scene text detection may not be adequate to achieve better results for marathon and sports video. Thus, detecting text of marathon runners and sports players in video still remains as an unsolved problem.

This motivates us to explore skin, face and torso information for improving text detection performance in this work. As mentioned in the above, the human body is a prominent source of information in video of marathon and sports. Thus, in this work, we propose a new method for detecting skin, face and torso in video to address the challenges of detecting text on marathon runners and sports players in video. In Fig. 1text detection results of the state-of-the-art natural scene text detection methods and the proposed method are shown and from the Fig.1 it is evident that images have different challenges. It can be seen that the marathon image contains text of different degradations. Similarly, the sports image contains text which suffer from poor quality due to the folding of non-rigid material of clothes, human actions and structures of body parts.

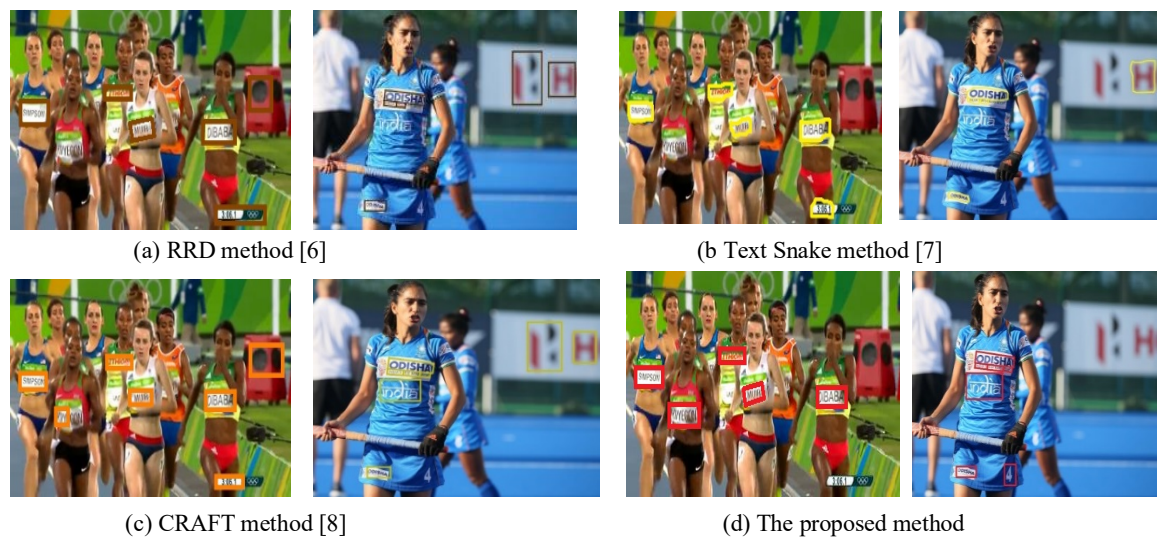

(a) RRD method [6]  (b Text Snake method [7]

(c) CRAFT method [8]  (d) The proposed method

Fig. 1. Text detection results of the proposed and existing methods of marathon runner and sports images. player.

To test the results of natural scene text detection methods, we use some recent methods, namely, Rotation-Sensitive Regression Detector (RRD) [6], which explores text location and rotation information, the TextSnake method [7], which explores flexible representation to handle arbitrary shapes, and the Character Region Awareness for Text Detection (CRAFT) method [8], which explores the affinity between characters for text detection in natural scene images. It is observed from Fig.1(a) that RRD misses texts in both the images, TextSnake misses the text in the sports image, while CRAFT misses' texts in the marathon image and the sports image. However, the proposed method is able to detect all the texts of the marathon runner and sports player including. Therefore, we can conclude that the methods developed for natural scene text images may not perform well for video images of marathon and sports. On the other hand, we can assert that the method which explores skin, face and torso can overcome the problems of text detection methods for marathon and sports video images. The main contributions of the proposed method are as follows. The proposed method explores gradient magnitude and directional coherence to determine the number of temporal frames in video. The new method based on a Bayesian classifier for skin pixel detection is used for segmenting torso regions of human bodies. In order to handle the complexities of detecting text of marathon runners and sports players, the proposed method combines torso region segmentation and deep learning-based text detection in a novel way, which is different from the existing methods.

## 2. Related Work

Since detecting texts from marathon runners and sports players in video is a part of text detection in natural scene images and video, we divide the related methods into four sub-categories as natural scene text detection [6-20], text detection in video [21-23], text tracking in video [24-27], and bib number/text detection in natural scene images [4, 5, 28].

### 2.1. Methods for Natural Scene Text Detection

Tian et al. [9] proposed to detect texts in natural images with a connectionist text proposal network. The method detects a text line in each sequence of fine-scale text proposals by convolutional feature maps. The method explores context information of images for successful text detection. The method is reliable for multi-scale and multi-lingual texts. However, the main objective of the method is to address challenges of natural scene text detection. Zhang et al. [10] proposed multi-oriented text detection with fully convolutional networks. The method considers both the local and global cues for localizing text lines in a coarse-fine procedure. The method focuses on multi-oriented, multi-language and multi-font situations, including low contrast images. Liu and Jin [11] proposed a method for fixing tight bounding boxes for text lines. The method proposes a new convolutional neural network-based method, which is called a deep matching prior network for text detection in natural scene images. The method works well for images affected by blur and low contrast. Endo et al. [12] proposed a scene text detection method, which is robust to orientation and non-contiguous components of characters. The method proposes a connected component-based scene text detection approach, which involves neighbor character search using a synthesizable descriptor for the non-contiguous component problem, and a region descriptor called a rotated bounding box descriptor for rotated characters. Tian et al. [13] proposed scene text detection under weak supervision. The method trains robust and accurate scene text detection by learning from an un-annotated dataset. For unsupervised learning, the light supervised model is applied to the un-annotated dataset to search for more character training samples, which are further combined with the small annotated dataset to retain a superior character detection model.

Deep learning has been used for text detection in natural scene images. Shi et al. [14] proposed detecting oriented texts in natural images by linking segments. The idea here is that each text component is decomposed into segments and links. A segment is an oriented text box that covers a part of a text line and a link is the connection of two adjacent segments, indicating that they belong to the same text line. To achieve this, the method explores fully convolutional neural networks. Zhou et al. [15] proposed an efficient and accurate scene text detector. The method predicts text lines without detecting words or characters based on deep learning concepts. The method works well for arbitrary orientations. Yao et al. [16] proposed scene text detection via holistic and multi-channel prediction. The method explores character information, text regions and their relationship for achieving better results. It uses a single fully convolutional neural network for text detection in images. He et al. [17] proposed accurate text localization in natural scene images with a cascaded convolutional text network. It proposes two networks, one for rough detection of text regions, and another for accurate text localization from large regions detected by the first network. Deng et al. [18] proposed a method for detecting scene texts via instance segmentation. The method divides a text into a number of instances, and then finds links among the instances to extract contextual information which helps to detect texts accurately. Liu et al. [19] proposed a method for detecting curved texts in the wild. The method proposes polygon-based curve text detection, which can directly detect texts without empirical combination. The proposed architecture allows context information instead of predicting points independently, which results in smooth text detection in images. Veit et al. [20] proposed text detection and recognition for Coco texts. The method focuses on solving multiple issues of text detection in natural scene images, such as curved texts, arbitrary orientation and backgrounds.

From the above discussions, it is noted that the methods address several challenges of text detection in natural scene images by exploring deep learning in a different way. However, the scopes of the methods are limited to text detection in natural scene images, where one can expect focused text information in images. If the methods are tested on marathon and sports images where the main focus is on persons, the performance of the methods may degrade due to inadequate information available for the texts.

## 2.2. Methods for Text Detection in Video

There are several methods for enhancing texts affected by low resolution and contrast, which are hindrances to video text detection methods. The methods generally use temporal information for magnifying low contrast such that the methods can overcome the problem of low contrast and low resolution. Hu et al. [21] proposed video text detection with text edges and a convolutional neural network. The method finds candidate text regions, and then segmented regions are sent to a convolutional neural network to generate a confidence map for each candidate text region. Candidate text regions are further refined and portioned into text lines by projection analysis. Mittal et al. [22] proposed rotation and script independent text detection from video frames using sub-pixel mapping. The method proposes sub-pixel mapping using the super resolution concept to enhance low contrast text information. Histogram-oriented moment features are extracted for connected components with the help of a support vector classifier for text and non-text identification. Furthermore, a recurrent neural network classifier has been used for character recognition. Shivakumara et al. [23] proposed a fractal-based multi-oriented text detection system for recognition in mobile video images. The method explores the properties of fractals such as self-similarity in the gradient domain for enhancing texts, and then the same fractals expansion is proposed in the wavelet domain for multi-sized and multi-font text detection. Optical flow properties of texts have been used for eliminating the false positives. Further, direction-guided boundary growing is proposed for text detection.

It is found from the above discussion that exploring temporal information for the pre-processing step helps to enhance the fine details in video images. In other words, the methods do not utilize spatio-temporal information to discriminate text and non-text pixels in images for text detection. This shows that the methods focus on one aspect, such as low contrast and low resolution for improving text detection performance. However, this idea may not work well for marathon and sports video because text is affected by several adverse factors, such as occluding text with the color of clothing, distortion due to material folding, human actions, structure of human body parts etc.

## 2.3. Methods for Text Tracking in Video

Unlike the methods discussed in the previous section, there are methods that explore spatio-temporal information for classifying text and non-text pixels by extracting spatial relationships between temporal frames. The same information has been used for tracing text in video. Shivakumara et al. [24] proposed a new technique for multi-oriented scene text line detection and tracking in video. The method explores gradient directional symmetry features at the component level for smoothing edge components. Then it uses spatial information given by Delaunay triangulation to obtain text candidates. Text candidates are merged to form text lines based on the nearest neighbor approach. Text lines are traced by matching subgraphs of triangulation. Temporal information is used for tracking texts after text line detection. Zuo et al. [25] proposed multi-strategy tracking-based text detection in video scenes. The method detects texts in frames and then proposes multi-strategy tracking based on spatio-temporal context learning and linear prediction. A rule-based method is proposed for tracking texts by finding the best match. However, it is not clear how to determine the number of frames for tracking texts. In addition, the method was not explored on Marathon or sports videos. Tian et al. [26] proposed a unified framework for tracking-based text detection and recognition from web videos. The method proposed a generic Bayesian-based framework for tracking texts in video. It is noted that the method explores temporal frames for enhancing text detection performance. In addition, the method considers only caption texts for tracking in video. However, it is unclear how the method performs for texts in Marathon and sports videos. Yang et al. [27] proposed tracking-based multi-orientation scene text detection. The method follows the same conventional way of detecting texts in individual frames, and then text tracking is performed based on a dynamic programming concept. The method explores temporal information for collecting evidence for accurate text detection in frames. The method is limited to scene texts in video but not bib number/text in Marathon and sports videos.

It is observed from the above review that the methods trace texts in video but it is not clear whether these methods can trace texts of marathon runners and sports players because the latter depends on person movements but not text movements. Therefore, the methods may not perform well for detecting text from marathon runners and sports players in video. In addition, the methods ignore determining the number of temporal frames for the text detection process.

## 2.4. The methods for Bib Number/Text Detection from Marathon Runner and Sports Player Images

There are a few methods available in the literature for bib number/text detection in marathon images, which explore face and torso information for text detection. Ami et al. [4] proposed racing bib number recognition.

The method uses face detection as the first step to find torsos, and then detects bib numbers in torso regions. Since its focus is to recognize bib numbers, the scope is limited to bib numbers in marathon images but not sports where we can see not only bib numbers but also other texts over human bodies. To overcome this limitation, Shivakumara et al. [5] proposed a multimodal approach for bib number/text detection and recognition in marathon images. In this method, grab cut was explored for torso segmentation with the help of HOG features and an SVM classifier. Then a HOG-based text detection step is used for detecting bib numbers as well as texts. However, it is noted that the method requires high quality images for successful results. Kamlesh et al. [28] proposed an approach for person re-identification with end-to-end scene text recognition. This method explores text detection and recognition for person re-identification in Marathon images. The method tunes the deep learning model, which was proposed for text detection in natural scene images, for detecting bib numbers in Marathon images. However, according to the description of the dataset, the method considers high-quality images having black texts in the foreground and pure colors in the background. These constraints are hard for achieving better results for detecting texts of marathon runners and sports players in video. Overall, it is noted that the methods target detecting bib numbers of marathon runners and texts of sports players. But the methods are not robust because the performance depends on either face detection results or segmentation of torso regions. When an image contains multiple persons and groups of people, face detection mat not work well. At the same time, segmenting torsos from images having multiple persons is not easy due to partial occlusion of human bodies. Additionally, the methods are confined to still images, which are high quality but not video which suffers from poor quality, low resolution and low contrast. Therefore, achieving better results for detecting text of marathon runners and sports players in video is still an open challenge for researchers in this domain.

Hence, in contrast to the existing approaches, this paper presents a new unified method for detecting bib number/text from marathon runners and sports players in video. The proposed method explores features, namely, gradient and directional coherence, to determine the number of temporal frames. Next, the proposed method explores skin features for face and torso detection, rather than detecting torso through face as in [4, 5]. This is because missing faces in Marathon and sports video are quite common due to occlusion. In addition, we explore Bayesian probability at the pixel-level of individual frames and at the component-level of temporal frames for detecting skin components. Furthermore, we propose to use a deep learning model for text detection from torso regions by modifying the loss function of the existing model [18]. Since the proposed method involves feature extraction for text candidate detection, skin and face detection for torso estimation as well as deep learning for text detection from torso regions, the proposed method is considered as a unified one. Thus, the proposed method is different from the existing methods [4, 5] for detecting text from marathon runners and sports players in video.

## 3. The Proposed Methodology

In this work, we consider Marathon and sports videos with temporal frames captured at the rate of 25-30 frames per second as the input. In other words, the proposed method works when video contains human bodies having text information. Therefore, it is rare to find videos without persons in this context, to the best of our knowledge, and hence this constraint may not be regarded as a vital issue. It is noted that background information (other than humans) in temporal frames usually has tiny movements, while foreground information (any part of a human or a whole human) stays at the same location for at least 10 frames [24]. To extract such observations, we were motivated by the method in [29] where the Gradient Magnitude (GM) and Directional Coherent (DC) features were proposed for detecting blurred regions. We propose to explore the same features for detecting candidate regions (foreground information) from temporal frames. The proposed method normalizes the values of GM and DC to the range of 0 and 1. Then the two images are fused by aggregating the maximum difference chosen from defined sliding windows over GM and DC images. The proposed method finds frame differences from the first fused frame with the successive fused frames, which results in an image containing stable pixels. Due to complex backgrounds and static pixels in the background, the differences may contain both background as well as foreground pixels. Therefore, for the difference in results, the proposed method performs k-means clustering with k=3, which outputs three clusters, namely, Max which contains high values, Min which contains low values, and Average which contains neither high nor low values. We believe the Max cluster represents edge pixels, the Min cluster represents background pixels, while the Average cluster represents skin pixels. As temporal frames pass, the frame differences result in huge changes in Max, Min and Average clusters. These changes are represented by statistical features, namely, mean, median and standard deviation of clusters, which give not only stable regions (candidate regions) but also help in determining the number of temporal frames and key frames, automatically.

The proposed method considers fused images of temporal frames for skin detection, where the images contain stable pixels including edge pixels of texts and skin pixels. It is true that the pixels, which represent skin, have a high degree of similarity compared to the other parts of the human body [30]. This cue motivated us to explore Bayesian probability by estimating a priori probabilities using pixels in Max and Average clusters, and the conditional probability using the number of repeated pixels in each defined window. This step classifies skin pixels in temporal frames. We estimate the same Bayesian probability at the temporal frame level by considering skin pixels in all the temporal frames, which results in a single image containing true skin components. In this way, the proposed method utilizes temporal frames for skin pixel detection.

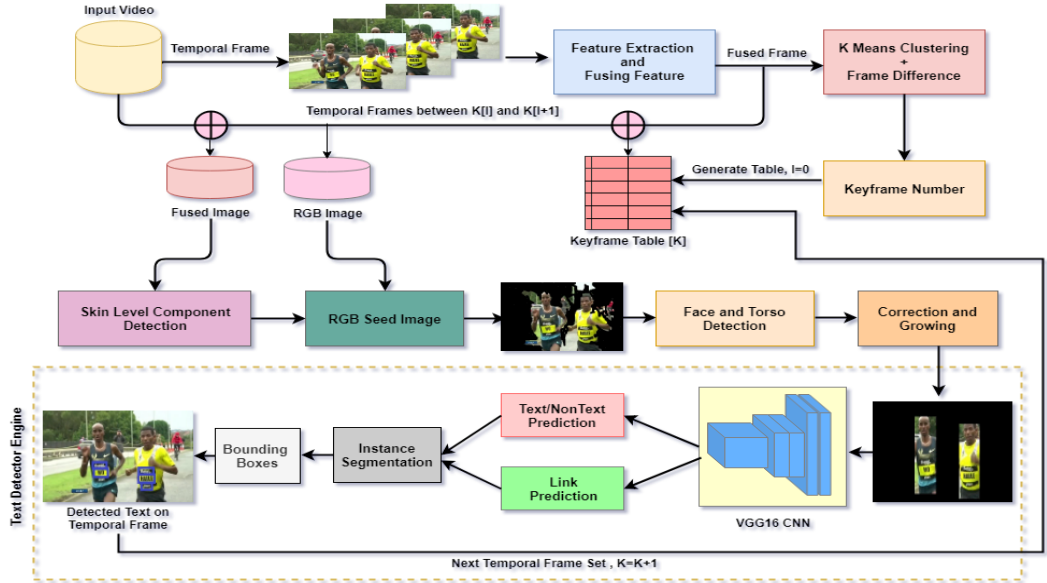

Fig.2. The overall framework of the proposed method.

It is noted that the presence of skin usually indicates the location of a face, shoulder, leg and/or torso. The proposed method combines all the skin components as one component based on nearest neighbor criteria, and considers it as the seed region for face and torso detection. For detecting faces from the seed region, the proposed approach uses a well-known method by Wang [31], which is available publicly and explores Haar wavelet features along with a cascaded classifier for face detection. According to the method [32], it is mentioned that the whole human body is structured, and the components such as face, upper and lower body have clear direct relationships in terms of geometrical measurement. In other words, with face location, one can locate torso and vice versa. Based on these observations, the proposed method detects and verifies face and torso in the seed region. The main advantage of this step is that if skin detection misses either face or torso, it is possible to restore the missing one. Furthermore, the proposed method explores a deep learning model, which involves text and non-text separation, grouping text components as a single element, for text detection from the torso region. The complete framework of the proposed method can be seen in Fig.2.

## 3.1. Magnitude and Directional Coherence Features for Candidate Region Detection

As mentioned in the previous section, for each pixel in the input image, the proposed method computes Gradient Magnitude (GM) and Directional Coherence (DC) as defined in Equation (1) and Equation (2) - (3), respectively, which outputs two images for the input image:

$$GM(i) = \log\left[\frac{1}{N_{P_i}} \sum_{j \in P_i} \left\{\sqrt{\frac{S_x^2(j) + S_y^2(j)}{2}}\right\}\right] \quad (1)$$

where $S_x$ and $S_y$ denote horizontal and vertical gradient operations, respectively, $P_i$ denotes the $i^{th}$ pixel position in the defined 3×3 window and $N_{P_i}$ denotes the number of pixels in the window. To widen the values between strong and weak edges, we perform a $log\,(*)$ operation for gradient magnitude values, and finally the values are normalized in the range of 0 and 1.

$$E(i) = \begin{bmatrix} \sum_{j \in p_i} S_x^{\,2}(j) & \sum_{j \in p_i} S_x(j).S_y(j) \\ \sum_{j \in p_i} S_y(j).S_x(j) & \sum_{j \in p_i} S_y^{\,2}(j) \end{bmatrix} \quad (2)$$

where $P_i$ is the same as GM, and $E(i)$ denotes the covariance between convoluted Sx and Sy, which outputs two Eigen values ($\lambda 1$ and $\lambda 2$) for the $i^{th}$ pixel. Here, the values of $\lambda_1$ and $\lambda_2$ indicate the degrees of anisotropy of the gradient distributions in the window. As noted in [29] where the dominant direction is useful in differentiating focused and defocused edges, we estimate Directional Coherence (DC) as defined in Equation (3) for the covariance values given by Equation (2).

$$DC(i) = \left(\frac{\lambda_1 - \lambda_2}{\lambda_1 + \lambda_2}\right)^2 \quad (3)$$

where $\lambda_1$ and $\lambda_2$ are computed from eigenvalue decomposition of matrix E($i$). It can be understood that DC is close to 1 if the edges in a patch are aligned, and DC is close to 0 if it has no dominant direction. The effect of GM and DC can be seen in Fig.3 for the input sports video image in Fig.3(a), where one notices that strong edges are sharpened in both GM and DC as shown in Fig.3(b) and Fig.3(c), respectively.

It is evident from Fig.3(b) and Fig.3(c) that pixels which have high contrast represent edge information. Since the proposed work is interested in extracting not only edge information but also other information such as face, torso regions and skin pixels, we propose to perform a difference operation over GM and DC images to enhance the pixel values of face, torso and skin regions. The difference operation is defined as finding the difference between the maximum and minimum element chosen from the sliding window of 5×5 pixels. For each window operation, the proposed method replaces the center pixel of the window by the difference value. This result in two difference images, namely, $GM_{diff}$ and $DC_{diff}$, of the same size as the GM and DC as defined in Equation (4) and Equation (5), respectively. This operation enhances not only edge pixels but also its neighbor pixels. To take the advantage of $GM_{diff}$ and $DC_{diff}$, the proposed method combines the difference images of GM and DC as defined in Equation (6) by adding the values in $GM_{diff}$ and $DC_{diff}$. The results are a fused image as shown in Fig.3(d), where one can see the pixels that represent edges as well as enhanced neighbor information (sharpened).

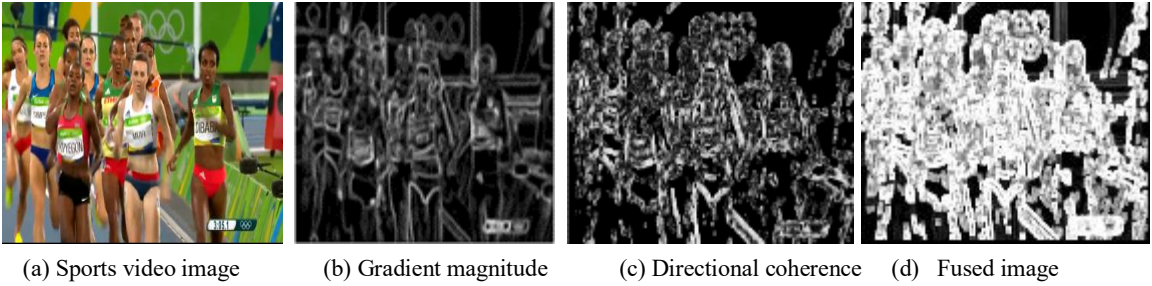

(a) Sports video image  (b) Gradient magnitude  (c) Directional coherence  (d) Fused image

Fig.3. Results after fusing gradient magnitude and directional coherence images.

$$GM_{Diff}(i) = \max(GM) - \min(GM) \qquad (4)$$
$$DC_{Diff}(i) = \max(DC) - \min(DC) \qquad (5)$$
$$Fuse(i) = GM_{diff}(i) + DC_{diff}(i) \qquad (6)$$

To determine the number of temporal frames for each key frame, the proposed method finds the difference between the first fused image with successive fused temporal frames. It is true that the pixels that represent foreground information such as objects and faces stay at the same location for at least 10 frames. As a result, the difference between the first fused temporal frames and successive fused temporal frames increases as the temporal frames passes. When there is a change in both foreground and background, one can expect a huge difference between the first fused temporal frame and the other fused temporal frames. At this point, the proposed method terminates the process of finding the difference and considers the frame number at which the process terminates as the actual number of temporal frames for further processing.

It is noted from Fig.3(d) that the fused image contains high, low and neither high nor low (average) values, which represent edges, background, and other information such as skin or torso regions, respectively. This observation leads us to apply K-means clustering with K=3 on each of the difference results to fix the condition automatically for determining the number of temporal frames. K-means clustering with K=3 classifies the values in the difference image into three clusters, namely, Max, Average and Min, as shown in Fig.4(a)-Fig.4(c), respectively. It is observed from Fig.4(a)-Fig.4(c) that most of the pixels which represent edges and objects are classified in to the Max cluster, skin pixels are classified either into the Max cluster or the Average cluster, while background pixels are classified into the Min cluster. It is expected that the presence of prominent edges will usually be the location of a face, objects, human body etc., but when there is a plain region, we can expect background information. This cue indicates that the Max cluster contains edges of face, human body, torso, leg, etc., the Min cluster contains background information, while the Average cluster contains neither edge nor background values, which includes skin pixels because pixels that represent skin have neither high value like edges nor low values like background pixels.

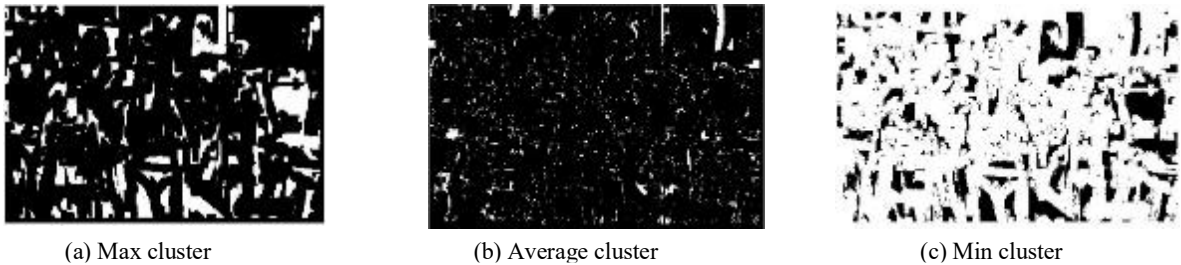

(a) Max cluster  (b) Average cluster  (c) Min cluster

Fig.4. Classifying pixels of edges, regions and background information into three clusters, Max, Average and Min clusters, respectively.

It is understandable from the illustration shown in Fig.5, where one can see that for skin patches of different colors of persons marked in (a) and (b), the standard deviation of intensity values for all the patches is almost zero as shown in Fig.5(c). Therefore, we use clusters given by K-means clustering to study the changes in the frame difference results. It is expected that the change mentioned in the above should be reflected in the clusters. The proposed method extracts the values in the fused images for corresponding to the pixels in the Max, Min and Average clusters. It calculates the standard deviation for Max and Avg clusters and the median for Min cluster to analyze the changes in frame differences. The histograms for standard deviation and median values of the respective cluster values are plotted as shown in Fig.6(a)-Fig.6(c), respectively, where we can see that there is a sudden change in the fused values on the Y axis either downward or upward as the number of temporal frames passes on the X axis. The sudden change can be any two out of three operations, which is considered as the terminating criteria as defined in Equation (7)-Equation (9). Fig.6 shows that the terminating process Fig.6(a) and Fig.6(c) exhibit sudden changes on the Y axis as the number of temporal frames increases on the X axis. In this way, the above-mentioned three operations are used for determining the number of temporal frames in this work. The same process also helps us to find the key frames in video. In this way, the proposed method finds the number of temporal frames using spatio-temporal information and key frames.

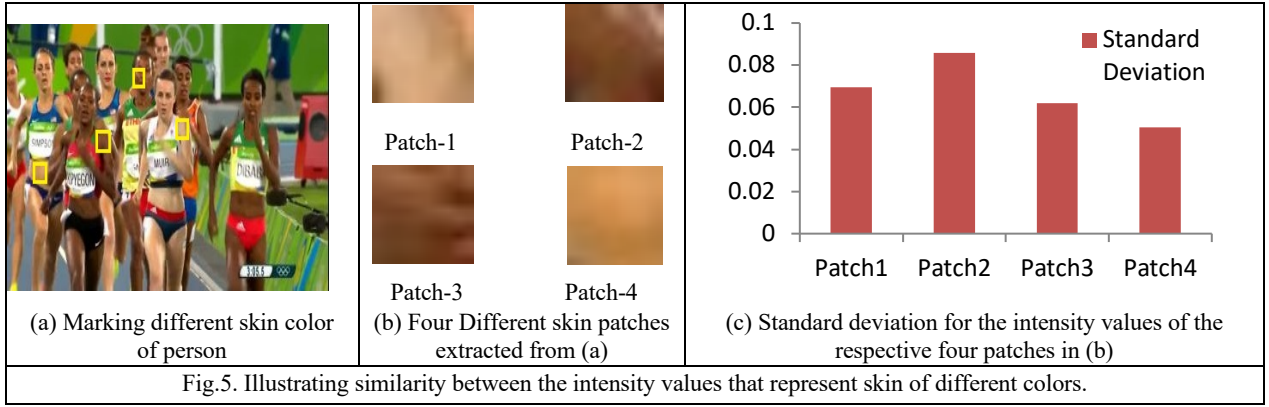

(a) Marking different skin color of person

(b) Four Different skin patches extracted from (a)

(c) Standard deviation for the intensity values of the respective four patches in (b)

Fig.5. Illustrating similarity between the intensity values that represent skin of different colors.

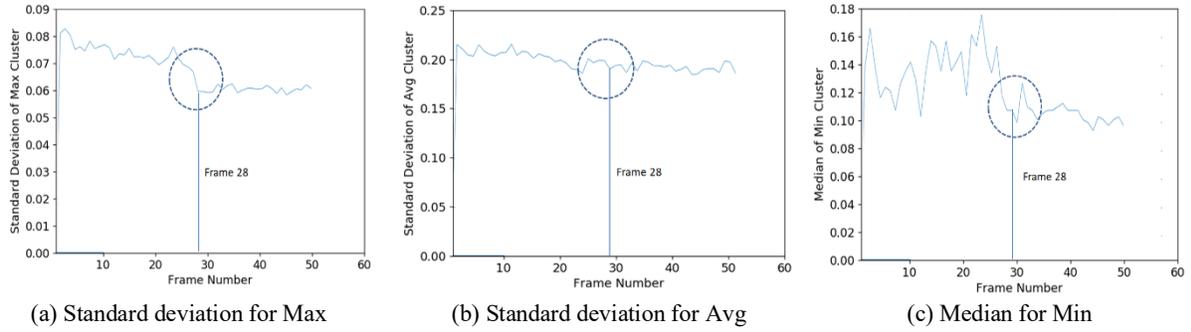

(a) Standard deviation for Max

(b) Standard deviation for Avg

(c) Median for Min

Fig.6. Illustration for determining the number of temporal frames automatically. In (a) and (b), the X axis (horizontal) denotes the number of temporal frames and the Y axis denotes the standard deviation for the values in the frame difference results. In (c), the Y axis denotes median values of the frame difference results.

$$C_{avg} = Normalized\ Max/Min/Avg\ Cluster\ of\ KMeans_{k=3}(Frame\_Difference)$$

$$[SD_{max/avg}]_f = \left[\sqrt{\frac{\sum_{i=1}^{N}(x_i - \breve{x})^2}{N-1}}\right]_f\ where\ x_i \in C, f \in Keyframe\ Number \quad (7)$$

$$[Median_{min}]_f = Median\ value\ of\ the\ Min\ Cluster \quad (8)$$

$$Keyframe = \min\left(round\left(\frac{x_z - \hat{x}_z}{\sigma_z}\right)\right) where\ z \in SD_{max/avg}/Median_{min} \quad (9)$$

Where $X_z$ are the values from $SD_{max/avg}/Median_{min}$, $\hat{x}_z$ is the mean of all the values in $SD_{max/avg}/Median_{min}$, and $\sigma_z$ is the standard deviation of the values in $SD_{max/avg}/Median_{min}$.

## 3.2. Bayesian Probability for Skin Component Detection using Temporal Information

Motivated by the method in [33] where a Bayesian approach has been used for classifying text and non-text pixels, we explore the same Bayesian probability for detecting skin pixels in frames. For the posterior probability, we need to estimate a priori probability and conditional probability for skin pixels. Since there is a rare chance of classifying skin pixels into a Min cluster, we ignore the pixels in the Min cluster and consider the pixels in the Max and Avg clusters as shown in Fig.7(a), which illustrates the color values corresponding to the pixels in the Max and Average clusters. The proposed method estimates priori probabilities for skin pixels by defining a sliding window of size 3×3 over the fused image corresponding to skin pixels in the Max cluster and the Avg cluster as defined in Equation (10), where the modes of Avg and Max clusters are considered. Similarly, the conditional probability is estimated as the mode of non-zero pixel values in the defined window divided by the number of pixels (fused values) in the window. Furthermore, the proposed method estimates the posteriori probability using the priori probability and conditional probability as defined in Equation (11), which classifies every pixel in the Max and Avg clusters as either skin pixels (white) or non-skin pixels as dark. The condition for the posteriori probability is greater than or equal to 0.50, which is more than half of the values for classifying skin and non-skin pixels. This process outputs two results for the respective Max and Avg clusters. Further, the proposed method concatenates these two results into a single one as shown in Fig.7(b), where we can see a few non-skin pixels are eliminated compared to the results in Fig.7(a). As a result, the proposed method detects skin pixels for each individual temporal frame.

$$P(skin) = \frac{MODE(Avg\,Cluster)}{MODE(Avg\,Cluster) + MODE(Max\,Cluster)} \text{ and } P(\overline{skin}) = 1 - P(skin) \quad (10)$$

$$P(skin\,|\,window) = \frac{P(window\,|\,skin).P(skin)}{P(window\,|\,skin).P(skin) + P(window\,|\,\overline{skin}).P(\overline{skin})} \quad (11)$$

The proposed method performs morphological operations for each pixel in Fig.7(b) to group the nearest neighbor pixels of one or two pixel distances, which results in skin components. When we look at the skin components across temporal frames, it can be noted that the pixel values (fused values) of skin components in all the temporal frames share a high degree of similarity. In addition, there is a chance that pixels of background objects may be misclassified as skin pixels. Therefore, we can use the above-mentioned steps for classifying pixels at the component and frame levels. For this, the proposed method calculates the mean of the pixel values of skin components of the respective temporal frames. For the mean values of skin components, the proposed method uses the same steps (as above) for selecting the actual skin components by eliminating false skin components as shown in Fig.7(c). This step results in actual components in one image out of all the temporal frames as shown in Fig.7(d), where it can be seen that most of the skin components are restored for the input image. Note that if this process misses a few skin pixels, it does not have a substantial effect on bib number/text detection because we can estimate the torso region where texts appears based on a few skin pixels of the face.

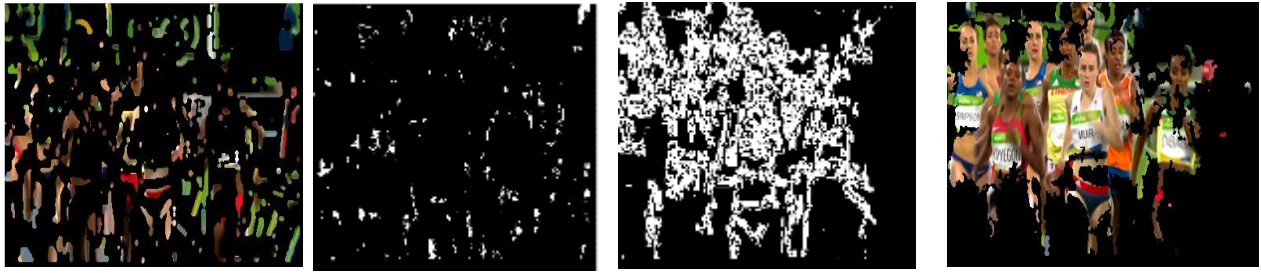

(a) Color-Max+Avg cluster　　(b) Result at the pixel level　　(c) Results at the component level　　(d) Skin components

Fig.7. Examples of skin detection results using Bayesian probability at the pixel and component levels.

### 3.3. Face and Torso Detection using Skin Components

In this step, color values corresponding to skin components given by Max and Avg clusters shown in Fig.7(d) are restored from the input images. The skin components with color values are used as the input. For the skin region in Fig.7(d), the proposed method uses a popular face detector called Viola Jones face detector [31], which considers Haar wavelet features and an Adaboost classifier for face detection. More details can be found in [31] about this face detector.

Due to complex backgrounds, the existing face detector [31] may not detect the correct facial region, rather it may lose a facial region or it detects extra non-facial regions. In this work, we prefer to use the method in [31] for face region detection rather than deep learning models because of the lack of ground truth and a large number of training samples. Therefore, in order to confirm the actual face region, the proposed method expands the face boundary and shrinks it pixel by pixel, iteratively. When the face detector loses face pixels, it results in a small region. In this case, the proposed method expands the boundary of the face iteratively and compares the confidence score given by the face detection method with the predefined samples. This process continues until the confidence score decreases after a few iterations. This indicates the boundary is covering non-facial areas and the region is considered as the actual face area. In the case that the face region is huge, the proposed method shrinks pixel by pixel iteratively as undertaken in the expansion process. This process terminates when the confidence score reaches the highest level and then starts decreasing after a few iterations. This indicates that the facial region is missing some vital face pixels and considers the region that gives the highest score as the actual face area.

Since skin components given by the previous step are the input for face and torso detection, there are four possible cases for torso region detection due to background variations in Marathon and sports videos. (i) skin components with face and torso regions, (ii) skin component with face and without torso, (iii) skin component with torso and without face, and (iv) skin component without face and torso regions. Out of the four cases, (i), (ii) and (iii) are ideal cases and case (iv) is beyond the scope of this research as the proposed method works based on skin detection. However, this case rarely occurs in practical scenarios.

**Case-(i) and (ii)**: In these cases, with the help of the face detector [31], the proposed method obtains the actual face area from which we can estimate height and width of the face region (head). As noted in [32] where it is stated that the height and width of a human body (torso) can be estimated using the height and width of the face (head) as defined in Equation (12) and Equation (13). This is possible because the whole human body is structured symmetrically and there is a relationship between all the parts of a human body according to [32]. Since the torso region is present below the head, we can estimate it using height and width of head information as defined in Equation (12) and Equation (13). This is illustrated in Fig.6(a), where we can see a torso region is detected by the proposed method using the face region (step by step). The values 7 and 2 mentioned in Equation (12) and Equation (13) are determined experimentally in this work by conducting experiments on 500 samples chosen randomly from all the datasets. According to our experiments, these values help us to detect not just exact torso regions, but the regions below the torso, such as the thighs of an individual. Therefore, the proposed method can detect texts on thighs also. This will be discussed in the Experimental Section.

$$Height\ Till\ Torso(T_h) = 7 * Height\ of\ Head(h) \quad (12)$$

$$Width\ of\ Torso\ (W) = 2 * Height\ of\ Head(h) \quad (13)$$

**Case-(iii)**: In this case, there is no requirement to detect facial regions using the torso region. This is because, generally, faces do not contain any text or bib numbers in Marathon and sports videos. To the best of our knowledge, torso or legs can contain text and bib numbers. Therefore, when a skin component provides the torso without a face region (as shown in one case in Fig.8(b)), it does not affect the text and bib number detection process. In this case, the proposed method considers the output of the skin component as the input, and then the measurements given by the method for the human body are checked to confirm whether the torso region covers a text region (or not). Since the proposed work focuses on bib number and text detection from the torso region, if the method based on human body measurements does not give the actual torso region, it does not have a major effect on the overall performance. Therefore, based on human body measurements, we fix the width and height of the torso region, and the same measurements are checked to confirm the torso region when there is no face in the skin component results as shown in the example in Fig.8(b).

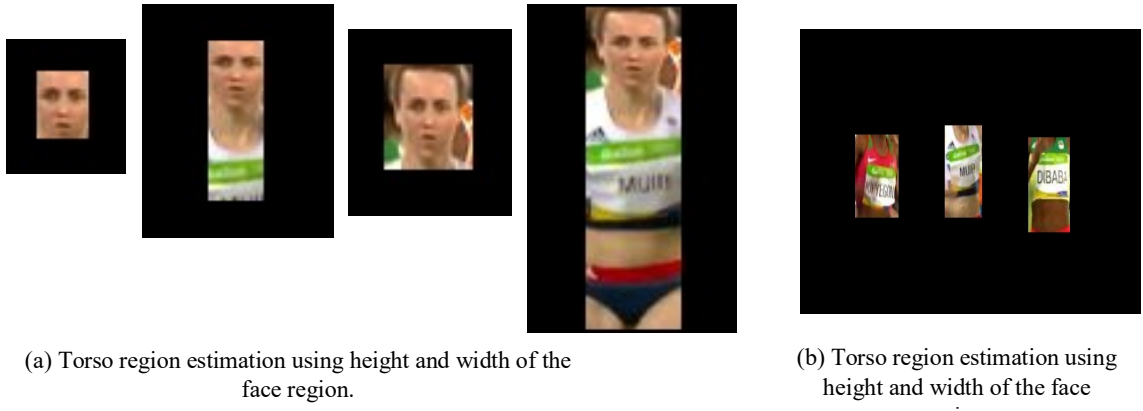

(a) Torso region estimation using height and width of the face region.

(b) Torso region estimation using height and width of the face

Fig.8. Examples of the results of Torso region prediction with and without face region.

### 3.4. Deep Learning Model for Text Detection from the Torso Region

Motivated by the method in [34] where the combination of feature extraction and deep learning provides stable results for complex issues, in this work, we explore the same combination for achieving better results for complex text detection in Marathon and sports video. Therefore, this section considers torso regions detected by the step presented in the previous section and a deep learning model for text detection from torso regions. As proposed in [18], we use the VGG16 backbone architecture, in which features of layer *fc6* and *fc7* are considered for feature extraction from torso regions or torsos recovered from face images. These features are then passed on to the next stage, where text and link classification takes place. Since the original implementation [18] is robust to any orientation based on its performance on the CTW1500 dataset, pixellink is selected as the base for detecting texts in sports videos. To achieve better and accurate results for Marathon and sports video, the proposed method uses the concept of instance segmentation [18] and linking text pixels [18] in this work. The linking algorithm extracts text locations directly from an instance segmentation result instead of from bounding box regression. The following conventions are used for linking text pixels. Pixels inside text bounding boxes are labeled as positive. If there is any overlapping pixel, only non-overlapped pixels are considered as positive, otherwise the pixels are considered as negative. To setup a link between positive pixels, the proposed method considers a seed pixel among all positive pixels. For a given seed pixel and one of its eight neighbors, if they belong to the same instance, the link between them is positive, otherwise it is a negative link. In this way, the proposed method uses instance segmentation and linking text pixels for text detection in marathon and sports video. More details can be found in [18]. The training loss is a weighted sum of loss on pixels and a loss on links as defined in Equation (14).

$$L = 2\alpha(t).L_{pixel} + (1 - \alpha(t)).L_{link} \qquad (14)$$

$$\text{where } \alpha(t) = \frac{1}{\log(10t+2)}$$

To handle the complexity of the proposed problem, in contrast to [18] where the method uses a constant weight in classifying positive pixels and linking of positive pixels, the proposed method aims to classify and link pixels dynamically instead of via fixed weights. This is due to the fact that linking positive pixels plays an important role in generating text instances. To achieve this, we adapt a new weighted loss function "L", where the weight function is $\alpha(t)$, which is a function of training epoch. The proposed adaptive weighted loss function shifts the weight of $L_{pixel}$ and $L_{link}$ accordingly, as training matures. It gives more weightage to $L_{link}$ as the training progresses because $L_{pixel}$ attains a discriminative power during training for classifying positive and negative text pixels. As a result, it is expected to put more weightage on linking positive text pixels since it is prone to errors in linking text pixels due to the pose of an athlete. The losses for positive and negative links are calculated separately, and on positive pixels, only as defined in Equation (15).

$$L_{pixel} = \frac{1}{(1+r)S} W L_{pixel\_CE} \qquad (15)$$

where $L_{pixel\_CE}$ is the Cross-Entropy loss matrix on link prediction. $W_{pos\_link}$ and $W_{neg\_link}$ are the weights of positive and negative links, respectively, which are derived from W. The detailed steps for the $k$-th neighbor of pixel $(i,j)$ are as follows.

$$L_{link\_pos} = W_{pos\_link} L_{link\_CE}$$
$$L_{link\_neg} = W_{neg\_link} L_{link_{CE}}$$
$$W_{pos\_link}(i,j,k) = W(i,j) * (Y_{link}(i,j,k) == 1)$$

$$W_{neg\_link}(i,j,k) = W(i,j) * (Y_{link}(i,j,k) == 1)$$

$$L_{link} = \frac{L_{link\_pos}}{rsum(W_{pos\_link})} + \frac{L_{link\_neg}}{rsum(W_{neg\_link})} \quad (16)$$

The notations used in Equation (16) are the same as are used in [17]. The details of training the proposed adapted deep learning model will be discussed in the Experimental Section. The block diagram of the above process can be seen in Fig.9, where we can see the flow for text detection in Marathon and sports video.

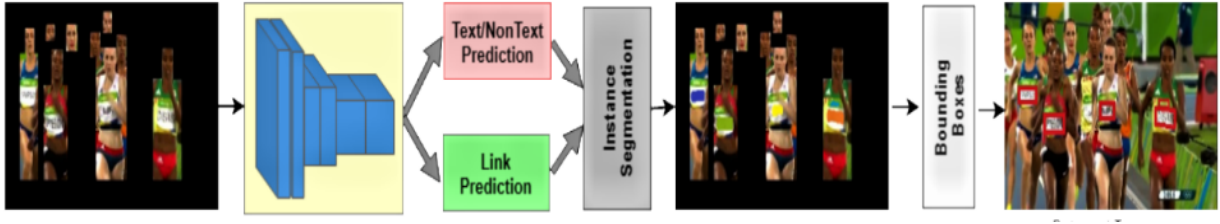

Fig.9. An architecture of the proposed adapted deep learning pixel liking model for text detection from torso region.

## 4. Experimental Results

For evaluating the proposed method, we created a complex dataset that includes videos of Marathons and sports events, namely soccer, tennis, cricket, etc. in contrast to the existing datasets [4, 5] that consider only Marathon videos. We use public resources, such as YouTube and the Internet, as well as our own collected videos for developing the video dataset. Each video clip considered in this dataset is captured for around 10 seconds at the rate of 25-30 frames per second. Since the data includes both marathon and other sports, we can expect challenges such as the effect of zoom-in or zoom-out cameras, the same view captured by different cameras at different angles, background changes, non-uniform illumination and different resolutions. In addition, video clips captured from sports usually contain both texts (scene text) and bib numbers especially for athletes; and video clips captured from Marathons contain bib numbers prominently. This variation in video makes the dataset more challenging. Similarly, in order to show the effectiveness of the proposed method, we consider three standard datasets that contains, Marathon video, namely, MMM [5], RBNR [4] and R-ID [28] where most of the images contain bib numbers. In the same way, to test the objectiveness and generic nature of the proposed method, we also collected images containing the human body with text from benchmark datasets of natural scene images, namely, CTW1500 [19] and MS-COCO Text [20], which provide images of different sports. Since the datasets are created for text detection in natural scene images, the images are much more complex compared to the other datasets. The details of the six datasets are reported in Table 1, where it can be noted that the total of 9115videos are used for experimentation. Sample frames of video clips for each datasets are shown in Fig.10(a)-Fig.10(d), respectively, where it can be noted that the frames from sports video clips can have text, bib numbers as well as scene texts. In addition, Table 1 shows that the proposed dataset includes the images of low-resolution imagery starting from 150×150 compared to other standard datasets. Therefore, the proposed dataset is challenging.

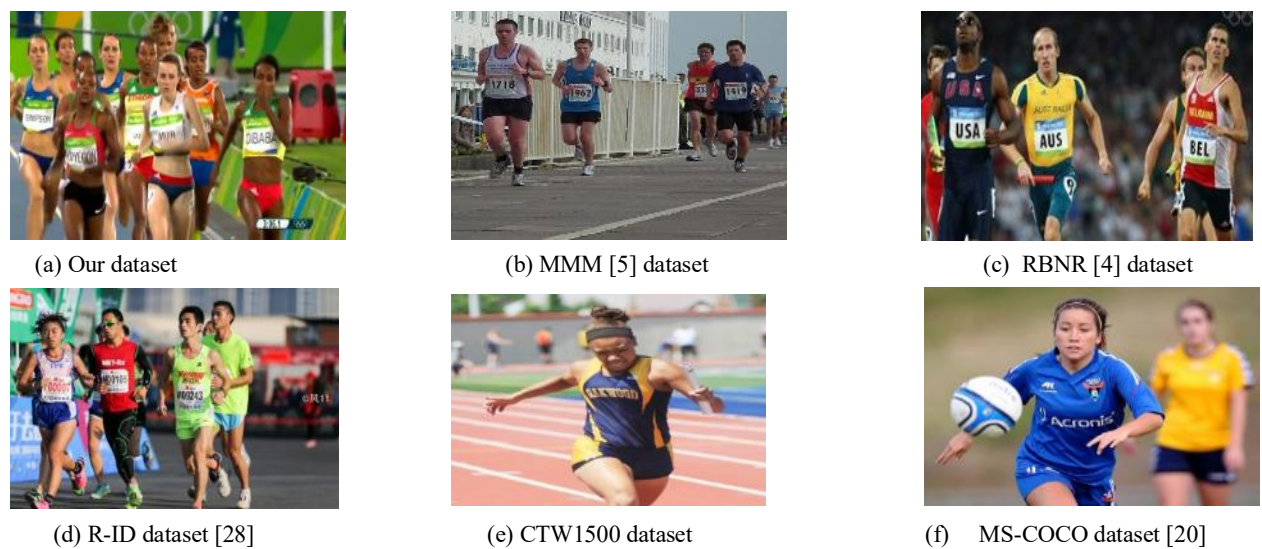

(a) Our dataset  (b) MMM [5] dataset  (c) RBNR [4] dataset

(d) R-ID dataset [28]  (e) CTW1500 dataset  (f) MS-COCO dataset [20]

Fig.10. Example of some sample frames from different datasets including our dataset.

Table 1. Details of the datasets used for experimentation in this work

| Datasets | Total Number of Videos | Resolution of Video | |
|---|---|---|---|
| | | Min | Max |
| Our Data-Sports and Marathon | 44 | 150 ×150 | 1280 × 720 |
| Shivakumara et al-Marathon [5] | 50 | 318 × 479 | 2939 × 1959 |
| RBNR Data-Marathon [4] | 50 | 342 x 479 | 1260 x 850 |
| Re-ID [28] | 8706 | 300 x 300 | 1200 x 800 |
| CTW1500 [19] | 15 | 620 x 437 | 1728 x 2592 |
| MS-COCO Text [20] | ~250 | 401 x 375 | 640 x 640 |
| Total | | 9115 | |

To measure the performance of the proposed method, we use standard measures, namely, Recall (R), Precision (P) and F-Measure (F) as defined in [4, 5]. F-measure is calculated using recall and precision as defined in Equation (17), where the value of $\alpha$ is fixed at 0.5 according to the instructions given in [4, 5]. More details on the definitions and formulae for recall, precision can be found in [4, 5].

$$F = \frac{P.R}{\alpha.R + (1-\alpha).P} \qquad (17)$$

To show the usefulness of the proposed method, we implement the following four state-of-the-art methods in order to conduct a comparative study: the first is Ami et al.'s method [4], which uses the combination of face, and torso for bib number detection from Marathon video frames. Shivakumara et al.'s method [5] uses face detection, torso detection and text detection for bib number detection in frames extracted from Marathon video. Kamlesh et al.'s [28] method uses text detection methods from natural scene images for bib number detection in Marathon images for person re-identification (Re-ID).Since the use of deep learning is popular for text detection in video frames and natural scene images, and has the ability to solve complex issues such as small fonts, font size, low contrast, complex backgrounds and multi-orientation such as for texts in Marathon and sports video, we use its code available online for a comparative study in this work. For example, Zhou et al. [15] proposed an efficient and accurate scene text detector, which explores deep learning to tackle the issues of small fonts and different datasets, along with orientation issues. Liao et al. [6] proposed a Rotation-Sensitive Regression Detector (RRD) for detecting texts in natural scene images. Long et al. [7] proposed the TextSnake method, which focuses on a flexible representation for detecting texts in natural scene images. Baek et al. [8] proposed Character Region Awareness for Text Detection (CRAFT) in natural scene images. Note that the above methods are developed for individual images or video frames but not for video with temporal frames. For a fair comparison, we fed the key frames extracted by the proposed method for text detection by the existing methods. The three standard datasets, namely, MMM [5], RBNR [4],Re-ID [28], CTW500 [19] and MS-COCO Text [20] do not provide temporal frames. In this case, we created duplicate frames of 25-30 for each image (as in the proposed dataset) for conducting the comparative experiments. Note: since our dataset and the MMM dataset do not provide ground truth, we compute the measures manually. However, since the CTW1500, MSCOCO, RBNR and Re-ID datasets provide ground truth, the proposed method calculates measures automatically.

For text detection from torso regions, we propose an adaptive deep learning pixel linking model. The details of training and learning are as follows. We use the dataset of the ICDAR 2015 robust reading competition for fine tuning the proposed adaptive deep learning linking model. The architecture is optimized by Stochastic Gradient Descent (SGD) with momentum = 0.8 and weight decay as $6 \times 10^{-4}$. The learning rate chosen for training the architecture was 0.0001 for the first 20K iteration and 0.001 for the next 40K iterations. The proposed architecture is trained with 8706 annotated bib number images chosen from the dataset [28], and 500 randomly selected images from datasets in [4,5] to add variations to the bib numbers from different sports images. Apart from that, the architecture was tested with 1000 testing samples available from the datasets in [4,28]. In total, 10,206 images are used for learning, out of which, 7206 images are used for training, 2000 images are used for validation and 1000 images are used for testing. We follow the same set up for training and testing the existing methods [6, 7, 8, 15] on all the datasets in this work.

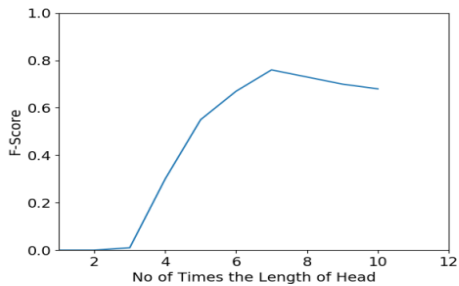

Fig.11. Determining the optimal values to estimate torso region using head height.

In Section 3.3, the proposed method uses height and width of the head for estimating torso regions by defining the relationship as being 7 times the height of the head and 2 times the width of the head to represent torso regions of the person. Value 7 is determined based on experimental results as shown in Fig.11, where the F-score of the torso region is calculated for different values given by proposed method. Fig.11 shows value 7 gives the best F-score and it drops as the value increases. Therefore, value 7 is considered as an optimal value for all the experiments in this work. In the same way, the proposed method finds value 2 for width of the torso region. For these experiments, we consider 500 sample images chosen randomly from all the datasets, and the proposed architecture was used for generating the F-score of the text detection.

As mentioned above, for the datasets MMM, RBNR, Re-ID, CTW1500 and MS-COCO, we create duplicate frames for each input image to make video for experimentation in this work. The proposed method finds frame differences for determining the number of temporal frames, and uses the same number of temporal frames for skin component detection using a Bayesian classifier. In these cases, when the input provides video of duplicate frames, the above process terminates because the process does not find any changes in the content of duplicate frames. It is evident from the results shown in Fig.12, where for the duplicate frames in (a), the steps of skin detection, face detection and torso detection give the same results for the both duplicate frames as shown in Fig.12(b)-(d), respectively. Therefore, we can conclude that creating duplicate frames may not have a substantial effect on the overall performance of the proposed method.

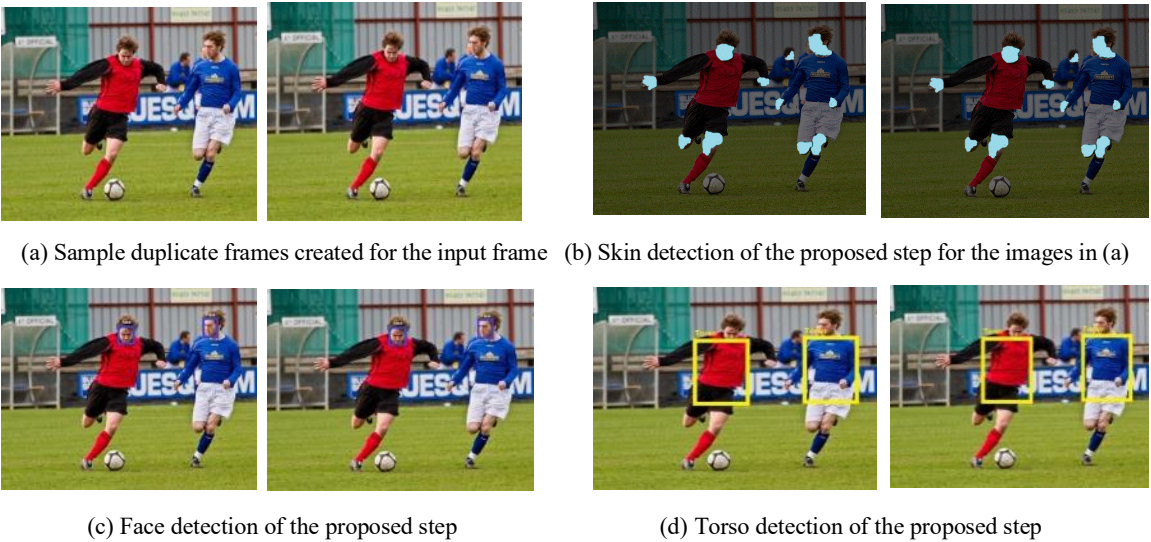

(a) Sample duplicate frames created for the input frame  (b) Skin detection of the proposed step for the images in (a)

(c) Face detection of the proposed step  (d) Torso detection of the proposed step

Fig.12. The performance of the steps of the proposed method when creating duplicate frames.

### 4.1. Evaluating Key steps of the Proposed Method

The proposed method consists of three key steps, which are: (i) key frame detection, (ii) skin, face, torso detection, and (iii) text/bib number detection in sports and Marathon video. To analyze the contribution of each key step, we evaluate each of these steps by calculating standard measures, such as recall, precision and F-measure. The qualitative results of each step are shown respectively in Fig.13(a)-Fig.13(c) for key frame, skin and face detection on our dataset. For different sample video clips, the proposed method successfully detects key frames as shown in Fig.13(a), where we can see that each frame represents a different video clip. For the same key frame, the proposed method detects skin properly as shown in Fig.13(b), where it is seen that the proposed method detects almost all the skin pixels for different situations. For face detection, since we provide skin detection results as the input toawell-known face detector, and deploy the iterative process that involves shrinking and expansion for the seed face area to accurately detect faces, the face detector provides successful detection results under different situations as shown in Fig.13(c), including for images in crowded situations. Quantitative results of the above three steps on our dataset are reported in Table 2, where it is noted that all the three steps contribute almost equally and score almost all consistent results for our challenging dataset.

Table 2. Performance of the key steps of the proposed method on our dataset

| Key Steps | Precision | Recall | F-Measure |
|---|---|---|---|
| Key Frame Detection | 0.86 | 0.91 | 0.88 |
| Skin Detection | 0.80 | 0.78 | 0.79 |
| Face Detection | 0.86 | 0.83 | 0.84 |

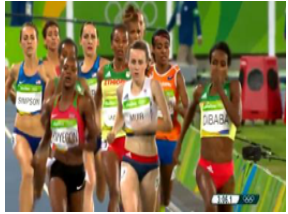　　　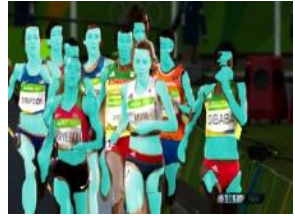　　　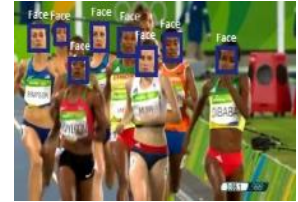

(a) Key frames detection　　　(b) Skin detection　　　(c) Face detection

Fig.13. Qualitative results of the key steps, key frame detection, skin detection and face detection of the proposed method on our dataset.

## 4.2. Evaluating Torso Detection

Qualitative results of the proposed and two existing methods, namely, Shivakumara et.al's method [5] and Ami et al.'s method [4] for torso detection on different datasets, namely, our dataset, the MMM dataset [5], the RBNR [4], the Re-ID [28], CTW1500 [19] and MS-COCO [20] datasets, are shown in Fig.14(a)-(c) to Fig.19(a)-(c), respectively. Since Ami et al.'s and Shivakumara et al.'s methods used torso detection steps for bib number detection in Marathon video frames, we use the same methods for a comparative study with the proposed method. It is observed from Fig.14(a)-(c) to Fig.19(a)-(c) that the proposed method works well for different situations compared to the existing methods. The reason for the poor results of the existing methods is that the methods are limited to bib number detection in Marathon images, and they use face detection for torso detection without temporal frames. On the other hand, the proposed method considers skin pixels detected by Bayesian probability with temporal frames as the input for face and torso detection. In addition, the proposed method uses the relationship between face and torso regions in terms of geometrical properties for torso detection with and without the face in contrast to the existing methods, which either assume fixed regions or use segmentation method.

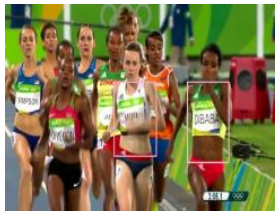　　　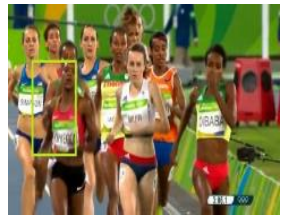　　　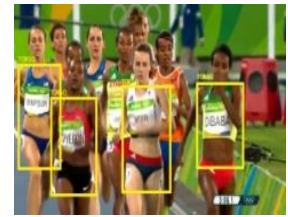

(a) MMM [5]　　　(b) RBNR [4]　　　(c) The proposed method

Fig.14. Qualitative results of the proposed and existing methods for torso detection on our dataset.

Quantitative results of the proposed and existing methods are reported in Table 3, where it can be noted that the proposed method provides the best results in terms of precision, recall and F-measure for the RBNR [4] and Shivakumara et al.'s datasets. The proposed method is also the best at precision for our dataset. Similarly, the proposed method achieves the best recall for the CTW500 and MS-COCO datasets, and the best recall and precision for Re-ID dataset. However, Shivakumara et al.'s method scores the best recall and F-measure for our dataset and the worst at precision on our dataset compared to the proposed method. This shows that the proposed method detects the torso accurately without many false positives compared to the existing methods. The reasons are the same as discussed above. Table 3 shows that most of the methods including the proposed method score the best results for RBNR and the Re-ID datasets compared to other datasets. Therefore, we can assert that the RBNR and Re-ID datasets are not as challenging to the extent that they are compared to other datasets including our dataset.

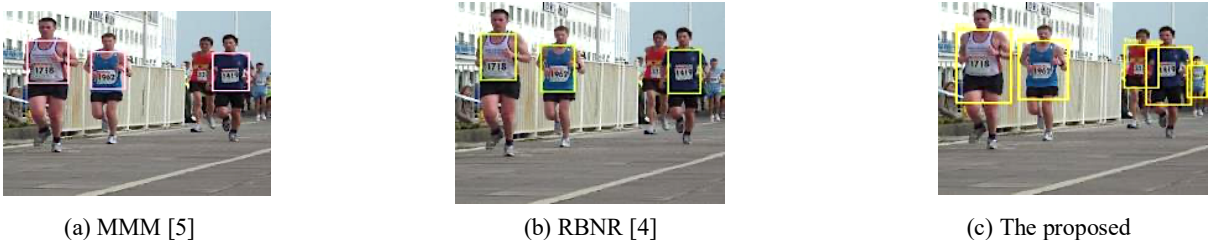

(a) MMM [5]  (b) RBNR [4]  (c) The proposed

Fig.15. Qualitative results of the proposed and existing methods for torso detection on the Marathon MMM [5] dataset.

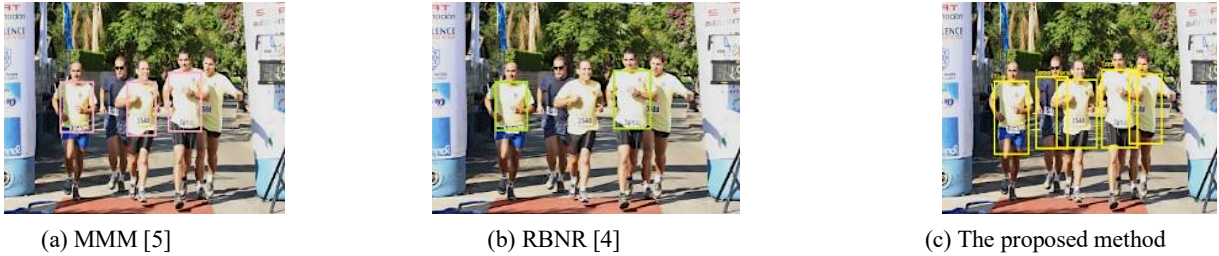

(a) MMM [5]  (b) RBNR [4]  (c) The proposed method

Fig.16. Qualitative results of the proposed and existing methods for torso detection on the Marathon RBNR dataset [4].

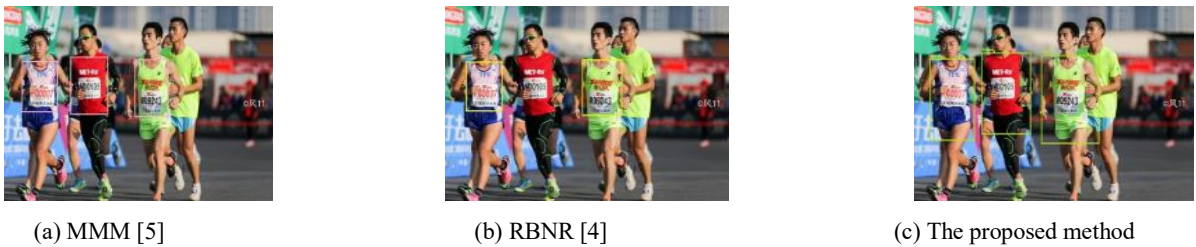

(a) MMM [5]  (b) RBNR [4]  (c) The proposed method

Fig.17. Qualitative results of the proposed and existing methods for torso detection on the R-ID dataset [28].

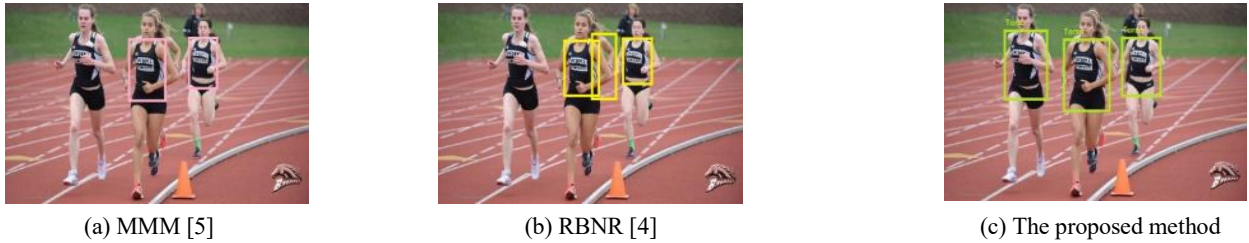

(a) MMM [5]　　　　　　　　　(b) RBNR [4]　　　　　　　　　(c) The proposed method

Fig.18. Qualitative results of torso detection of the proposed and existing methods for torso detection on the CTW1500 dataset [19].

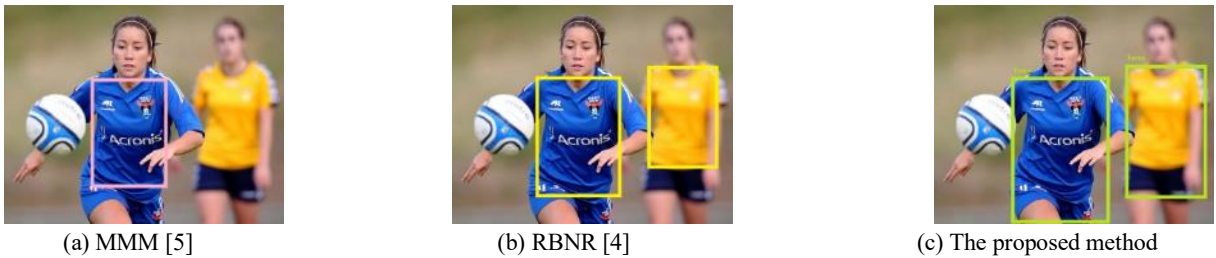

(a) MMM [5]　　　　　　　　　(b) RBNR [4]　　　　　　　　　(c) The proposed method

Fig.19. Qualitative results of the torso detection of the proposed and existing methods for torso detection on the MS-COCO dataset [20].

Table 3. Performance of the proposed and the existing methods for torso detection on different datasets.

| Methods | RBNR Dataset [4] | | | MMM Dataset [5] | | | CTW1500[19] | | | MSCOCO [20] | | | R-ID Dataset [28] | | | Our Dataset | | |
|---|---|---|---|---|---|---|---|---|---|---|---|---|---|---|---|---|---|---|
| | P | R | F | P | R | F | P | R | F | P | R | F | P | R | F | P | R | F |
| Proposed | **0.93** | **0.90** | **0.92** | **0.83** | **0.76** | **0.79** | 0.81 | **0.75** | 0.79 | 0.72 | **0.81** | **0.76** | **0.89** | **0.85** | **0.87** | **0.81** | 0.76 | 0.78 |
| RBNR [4] | 0.82 | 0.45 | 0.58 | 0.75 | 0.31 | 0.44 | 0.77 | 0.51 | 0.61 | 0.74 | 0.49 | 0.59 | 0.82 | 0.56 | 0.67 | 0.73 | 0.36 | 0.48 |
| MMM [5] | 0.88 | 0.86 | 0.87 | 0.78 | 0.76 | 0.77 | **0.86** | 0.74 | **0.80** | **0.76** | 0.73 | 0.74 | 0.83 | 0.84 | 0.84 | 0.77 | **0.80** | **0.79** |

## 4.3. Evaluating Text and Bib Number Detection

Qualitative results of the proposed and existing methods for text/bib number detection in Marathon and sports videos on six datasets are shown in Fig.20(a)-(h) to Fig.25(a)-(h), respectively. It is observed from Fig.20-Fig.25 that the proposed method detects both text as well as bib numbers successfully. But the existing methods [6, 7, 8, 15, 28], including multimodal methods [4, 5], do not perform well across all the datasets. The key reason is the features proposed do not have the ability to handle adverse situations unlike the proposed method. For instance, multimodal methods suffer from poor face detection as discussed in the above. The deep learning-based methods face issues of obtaining optimal parameters for achieving better results across all the datasets. This is understandable because the main aim of the existing methods is not text detection in both Marathon and sports datasets. For instance, the multimodal methods and the R-ID method were developed for bib number detection in Marathon images but not in sports video. Other methods are developed for text detection in natural scene images but not for text detection in Marathon and sports videos.

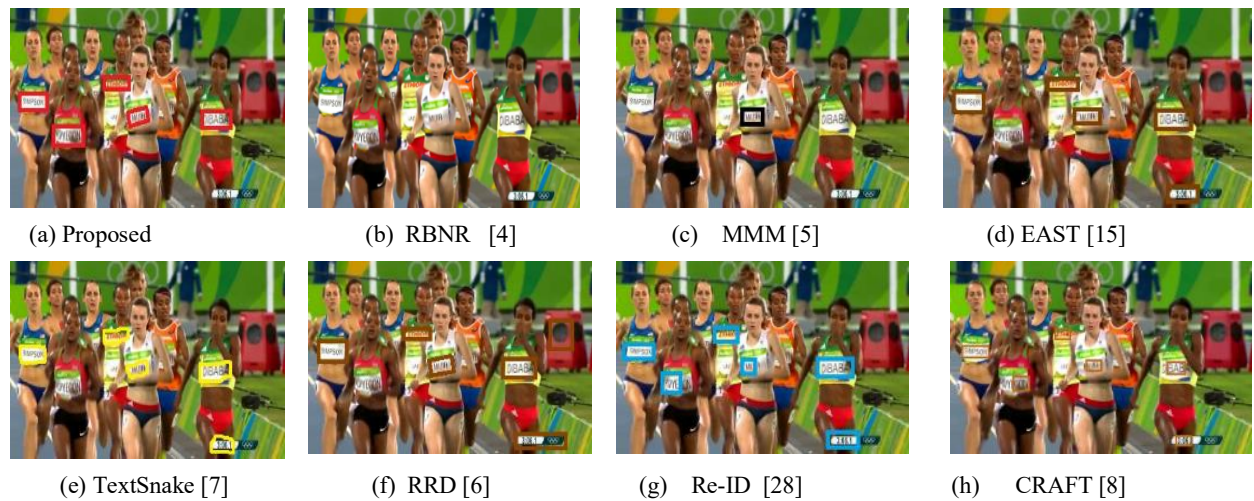

Fig.20. Qualitative text/bib number detection results of the proposed and existing methods on our dataset.

Quantitative results of the proposed and existing methods for six datasets are reported in Table 4, where it is noted that the proposed method is better at the F-measure than all the existing methods for all the six datasets. This shows that the proposed method is superior to the existing methods for detecting text of marathon runner and sports player in video. The main reason that the proposed method has achieved the best result is due to the exploration of skin, torso detection and the adaptive deep learning pixel linking model, which handles the adverse situations of different datasets. To understand the effectiveness of the existing methods that consider the full image as the input for text detection compared to the proposed method, which considers both torso region as input for text detection, we feed torso regions detected by the proposed method as the input for the existing deep learning-based methods. The results are reported in Table 4 as methods-torso. When we compare the results of image-full and image-torso, the results of image-torso are better than image-full. This is expected because torso detection reduces the complexity of the backgrounds in images. It is evident that torso detection improves text detection performance for Marathon and sports videos. Therefore, the results achieved by the text detection methods are close to the results of the proposed method. However, the results are not higher than the proposed method. It is observed from Table 4 that most of the methods, including the proposed method, score the better results for the RBNR and R-ID datasets compared to the other datasets. In the same way, when we compare the results of the proposed method on theCTW1500 and MS-COCO Text datasets with the results of other datasets, the proposed method scores the lowest results for CTW1500 and MS-COCO Text datasets. This is due to the unpredictable challenges (especially orientation) of texts in images compared to other datasets. This shows that these two datasets are still challenging and it is beyond the scope of the proposed work.

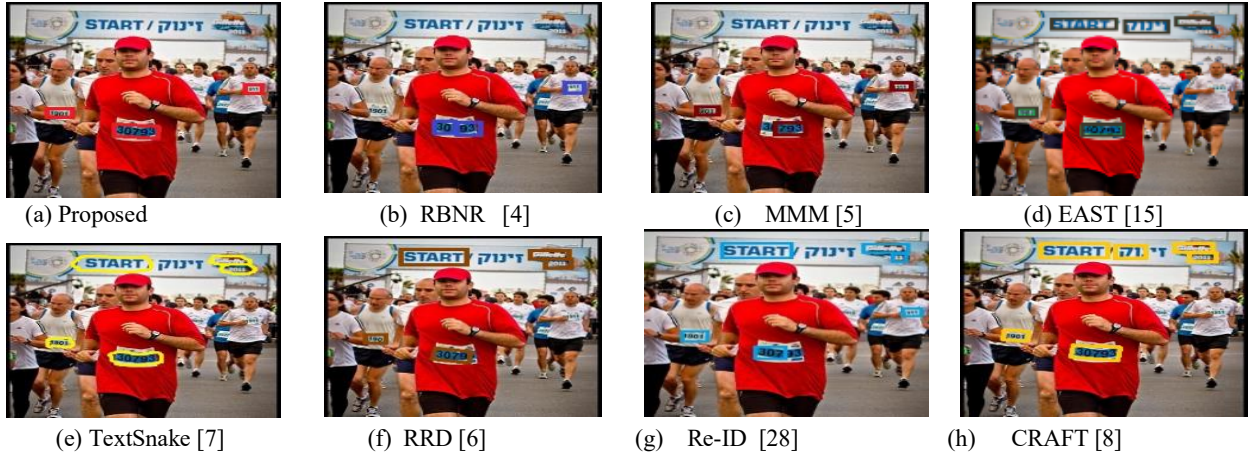

(a) Proposed     (b) RBNR [4]     (c) MMM [5]     (d) EAST [15]

(e) TextSnake [7]     (f) RRD [6]     (g) Re-ID [28]     (h) CRAFT [8]

Fig.21. Qualitative text/bib number detection results of the proposed and existing methods on the MMM dataset [5].

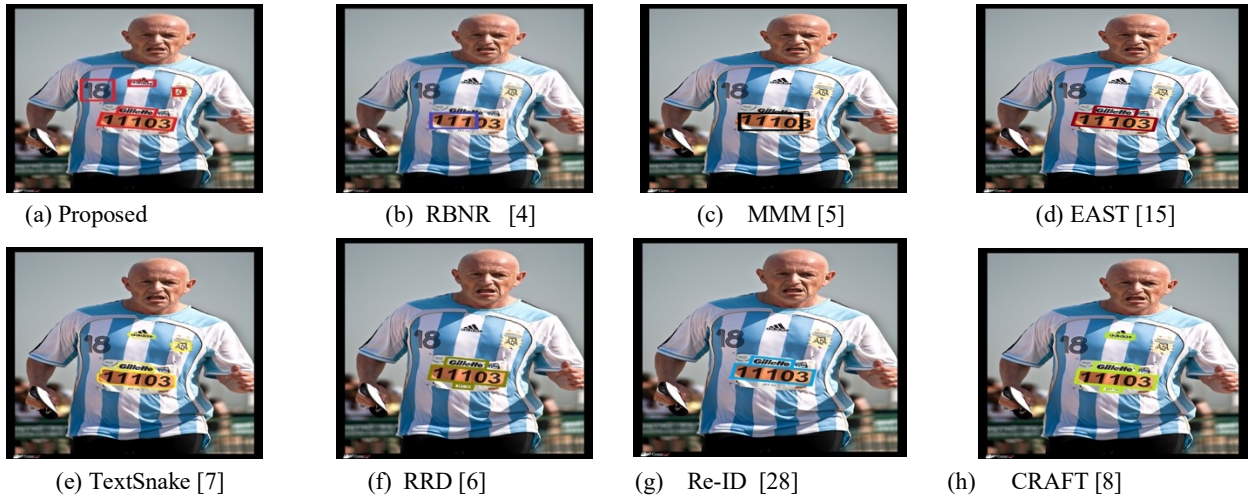

(a) Proposed     (b) RBNR [4]     (c) MMM [5]     (d) EAST [15]

(e) TextSnake [7]     (f) RRD [6]     (g) Re-ID [28]     (h) CRAFT [8]

Fig 22. Qualitative text/bib number detection results of the proposed and existing methods on the RBNR dataset. [4]

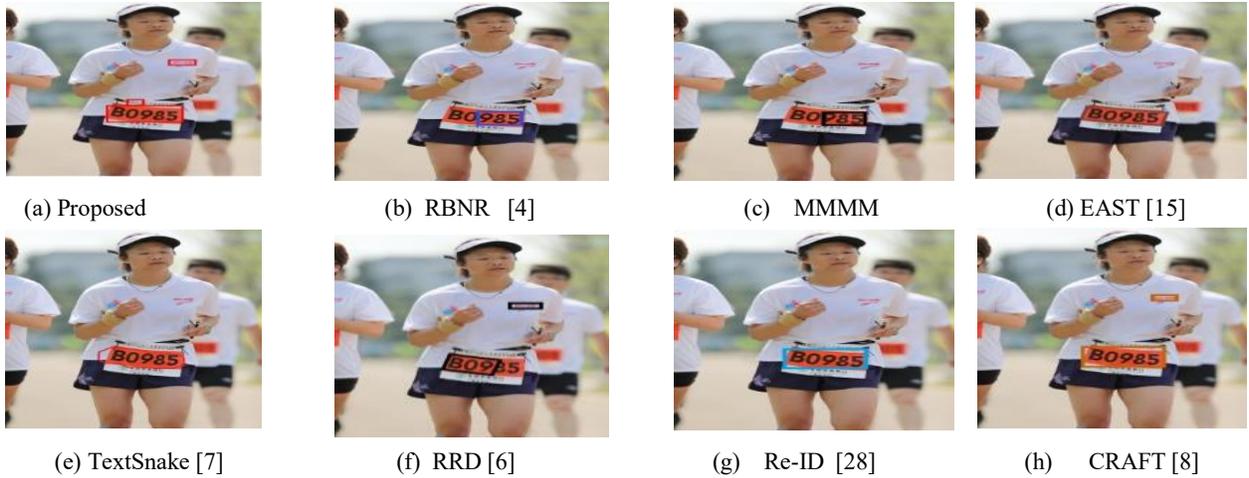

(a) Proposed     (b) RBNR [4]     (c) MMMM     (d) EAST [15]

(e) TextSnake [7]     (f) RRD [6]     (g) Re-ID [28]     (h) CRAFT [8]

Fig.23. Qualitative text/bib number detection results for the proposed and existing methods on the R-ID dataset [28].

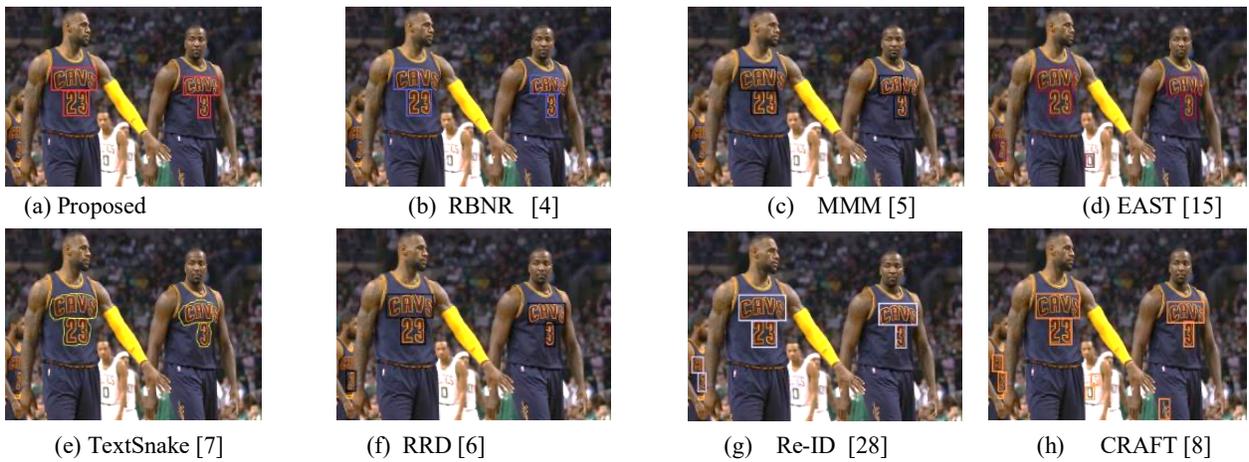

(a) Proposed     (b) RBNR [4]     (c) MMM [5]     (d) EAST [15]

(e) TextSnake [7]     (f) RRD [6]     (g) Re-ID [28]     (h) CRAFT [8]

Fig.24. Qualitative text/bib number detection results for the proposed and existing methods on the CTW1500 dataset [19].

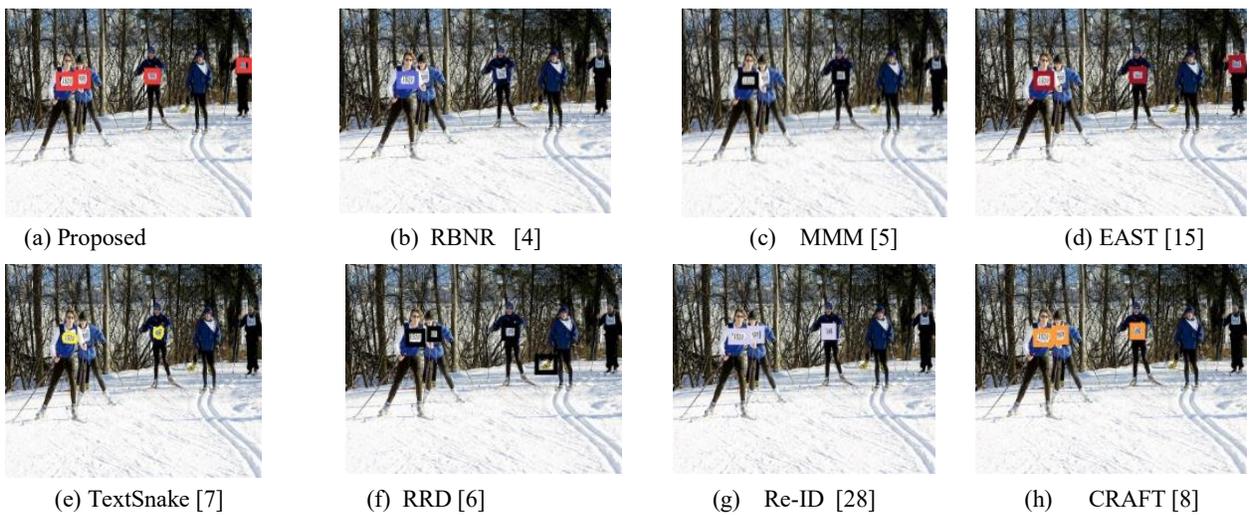

(a) Proposed     (b) RBNR [4]     (c) MMM [5]     (d) EAST [15]

(e) TextSnake [7]     (f) RRD [6]     (g) Re-ID [28]     (h) CRAFT [8]

Fig.25. Qualitative text/bib number detection results for the proposed and existing methods on the MS-COCO dataset [20].

Table 4. Performance of the proposed and the existing methods for text and bib number detection on different datasets

| Methods | RBNR Data [4] | | | MMM [5] | | | Re-ID Dataset [28] | | | CTW1500 [19] | | | MS-COCO [20] | | | Our Data | | |
|---|---|---|---|---|---|---|---|---|---|---|---|---|---|---|---|---|---|---|
| | P | R | F | P | R | F | P | R | F | P | R | F | P | R | F | P | R | F |
| **Proposed** | **0.74** | 0.76 | **0.75** | 0.78 | **0.80** | **0.79** | 0.89 | **0.92** | **0.90** | 0.73 | 0.69 | **0.71** | 0.38 | **0.37** | **0.37** | 0.81 | **0.83** | **0.82** |
| RBNR [4] | 0.53 | 0.40 | 0.45 | 0.37 | 0.38 | 0.38 | 0.67 | 0.63 | 0.65 | 0.30 | 0.33 | 0.31 | 0.20 | 0.16 | 0.18 | 0.46 | 0.50 | 0.48 |
| MMM [5] | 0.64 | **0.88** | 0.74 | 0.60 | 0.74 | 0.66 | 0.71 | 0.79 | 0.75 | 0.39 | 0.40 | 0.40 | 0.29 | 0.21 | 0.24 | 0.62 | 0.59 | 0.60 |
| EAST [15]-Full | 0.62 | 0.74 | 0.67 | 0.72 | 0.73 | 0.73 | 0.78 | 0.81 | 0.79 | 0.67 | 0.45 | 0.54 | 0.37 | 0.25 | 0.30 | 0.67 | 0.74 | 0.70 |
| RRD [6]-Full | 0.71 | 0.74 | 0.72 | 0.68 | 0.77 | 0.72 | 0.81 | 0.83 | 0.82 | 0.68 | 0.61 | 0.64 | 0.31 | 0.35 | 0.33 | 0.72 | 0.78 | 0.75 |
| TextSnake [7]-Full | 0.65 | 0.78 | 0.71 | 0.64 | 0.76 | 0.70 | 0.78 | 0.82 | 0.80 | 0.64 | 0.73 | 0.68 | 0.32 | 0.34 | 0.33 | 0.73 | 0.72 | 0.73 |
| Re-ID [28]-Full | 0.70 | 0.72 | 0.71 | 0.74 | 0.73 | 0.75 | 0.90 | 0.85 | 0.87 | 0.72 | 0.64 | 0.68 | 0.34 | 0.29 | 0.31 | 0.81 | 0.76 | 0.78 |
| CRAFT [8]-Full | 0.72 | 0.71 | 0.72 | 0.75 | 0.78 | 0.76 | 0.83 | 0.84 | 0.84 | 0.70 | 0.66 | 0.68 | 0.33 | 0.35 | 0.34 | 0.82 | 0.78 | 0.80 |
| **Detected Torso Images as input for Bib number/Text Detection** | | | | | | | | | | | | | | | | | | |
| EAST [13] – Torso | 0.64 | 0.79 | 0.71 | 0.74 | 0.75 | 0.75 | 0.81 | 0.82 | 0.81 | 0.69 | 0.52 | 0.59 | **0.40** | 0.29 | 0.34 | 0.72 | 0.79 | 0.75 |
| RRD [6] – Torso | 0.73 | 0.75 | 0.74 | 0.71 | 0.79 | 0.75 | 0.82 | 0.86 | 0.84 | 0.70 | 0.67 | 0.68 | 0.33 | 0.36 | 0.34 | 0.74 | 0.80 | 0.77 |
| TextSnake [7] – Torso | 0.67 | 0.79 | 0.73 | 0.68 | 0.77 | 0.72 | 0.80 | 0.87 | 0.83 | 0.66 | **0.76** | 0.71 | 0.34 | 0.37 | 0.35 | 0.76 | 0.79 | 0.77 |
| Re-ID [28] –Torso | 0.72 | 0.76 | 0.74 | **0.79** | 0.75 | 0.77 | **0.93** | 0.87 | 0.90 | **0.74** | 0.67 | 0.70 | 0.37 | 0.34 | 0.35 | 0.82 | 0.79 | 0.80 |
| CRAFT [8] –Torso | 0.74 | 0.74 | 0.74 | 0.77 | 0.80 | 0.78 | 0.85 | 0.88 | 0.85 | 0.72 | 0.69 | 0.70 | 0.36 | 0.37 | 0.36 | **0.83** | 0.80 | 0.81 |

In order to verify the conclusions drawn from the text detection experiments on the full images (input images) and torso regions detected by the proposed method, we use the same setup for conducting recognition experiments using the recent deep learning-based recognition methods. For these experiments, the following state-of-the-art recognition methods are used. Busta et al. [35] proposed anend-to-end trainable scene text localization and recognition framework. The method explores a single architecture for recognizing texts in natural scene images by considering the whole image as the input. Bartz et al. [36] proposed semi-supervised end-to-end scene text recognition. The aim of this method is to design a single architecture for both text detection and recognition similar to the method discussed in [35]. The methods explore deep learning models in a different way for addressing challenges of text recognition in natural scene images. Therefore, we use these two methods for conducting experiments in this work. Additionally, the methods presented in [4, 5] also include recognition steps along with bib number detection for marathon images. We use the above four methods for calculating recognition rates for texts in both the full and torso images from the marathon and sports scenarios. For calculating the recognition rate, we follow the instructions and measures presented in [4, 5]. More details about the definitions of measures and equations can be found in [4, 5]. The recognition results of the methods for full image and torso region for all the six datasets are reported in Table 5. It is observed from Table 5 that the methods score highly for torso regions compared to full images for all the datasets. This is expected because when the torso region is considered for recognition, the complexity of images are reduced as compared to the full one. The same conclusions are drawn from the text detection experiments. When we compare the results of [4, 5] with those of [35, 36], the latter is better than the former. This shows that deep learning-based methods have the ability to handle complex situations compared to those conventional methods, which use simple classifiers. It is noted from Table 5 that all the methods report low recognition rates for the MS-COCO dataset compared to other datasets, even for torso regions. This shows that texts in MS-COCO are complex in nature for achieving better recognition rates because this dataset is not created for text detection and recognition. Achieving better recognition rates for the MS-COCO dataset is beyond the scope of the proposed work. In summary, the conclusions drawn from the text detection experiments are the same as the conclusions drawn from the recognition experiments. Hence, we can conclude that torso region segmentation improves text detection and recognition performance significantly for marathon and sports videos. We can also infer that the proposed combination of torso region segmentation and text detection is better than existing methods for detecting text from marathon runners and sports players in video.

Table 5. Performance of the existing recognition methods for text and bib number recognition on different datasets

| Methods | RBNR Data [4] | | | MMM [5] | | | Re-ID Dataset [28] | | | CTW1500 [19] | | | MS-COCO [20] | | | Our Data | | |
|---|---|---|---|---|---|---|---|---|---|---|---|---|---|---|---|---|---|---|
| | P | R | F | P | R | F | P | R | F | P | R | F | P | R | F | P | R | F |
| RBNR [4] | 0.41 | 0.72 | 0.52 | 0.48 | 0.76 | 0.59 | 0.52 | 0.71 | 0.60 | 0.51 | 0.62 | 0.56 | 0.10 | 0.13 | 0.11 | 0.44 | 0.64 | 0.52 |
| MMM [5] | 0.39 | 0.69 | 0.50 | 0.35 | 0.62 | 0.45 | 0.47 | 0.60 | 0.53 | 0.49 | 0.65 | 0.56 | 0.12 | 0.14 | 0.13 | 0.56 | 0.71 | 0.63 |
| Deep Text Spotter [35]-Full | 0.64 | 0.78 | 0.70 | 0.72 | 0.70 | 0.71 | 0.74 | 0.76 | 0.75 | 0.68 | 0.73 | 0.70 | 0.12 | 0.15 | 0.13 | 0.79 | 0.75 | 0.77 |
| SEE [36]-Full | 0.69 | 0.73 | 0.71 | 0.75 | 0.76 | 0.76 | 0.81 | 0.83 | 0.82 | 0.74 | 0.76 | 0.75 | 0.14 | 0.20 | 0.16 | 0.82 | 0.79 | 0.80 |
| Detected text on Torso Images as input for Bib number/Text Recognition | | | | | | | | | | | | | | | | | | |
| Deep Text Spotter[35] – Torso | 0.71 | **0.80** | 0.76 | 0.75 | 0.79 | **0.82** | 0.84 | 0.83 | 0.85 | 0.71 | 0.78 | 0.74 | **0.16** | 0.20 | 0.18 | 0.80 | 0.78 | 0.79 |
| SEE [36] –Torso | **0.76** | 0.77 | **0.77** | **0.78** | **0.82** | 0.80 | **0.84** | **0.85** | **0.85** | **0.78** | **0.80** | **0.79** | **0.19** | **0.24** | **0.21** | **0.83** | **0.82** | **0.83** |

Despite the fact that the proposed method addresses issues in relation to marathon and sports videos, sometimes occlusions due to the crowd, severe blur due to text movements, non-uniform spacing between characters, and fonts which are too small due to the distance from the camera, as shown in Fig.26(a)-Fig.26(d), cause problems for the proposed method. Therefore, there is scope for further improvements and future work by proposing a super resolution to improve the image resolution and introducing context information for addressing the occlusion and blur-related problems

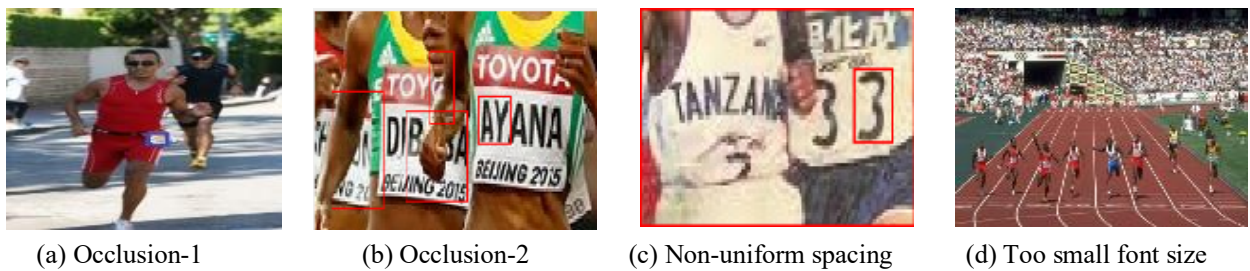

(a) Occlusion-1　　　　(b) Occlusion-2　　　　(c) Non-uniform spacing　　　　(d) Too small font size

Fig.26. Examples of erroneous results of the proposed method on our datasets.

## 5. Conclusions and Future Work

We have proposed a novel method for bib number/text detection in marathon and sports videos based on the combination of torso detection and a deep learning model. The proposed method takes advantage of skin, face and torso detection for reducing background complexity of the inputs to achieve better results. The proposed method explores gradient magnitude and directional coherent features in a new way for determining the number of temporal frames, which in turn gives key frame detection. The proposed method also uses Bayesian probability with temporal information in a different way for skin pixel detection. Skin pixels are used for face and torso detection by exploring the relationship between face and torso. Furthermore, we have proposed an adaptive deep learning pixel-linking model for text detection from the torso region. Experimental results on our own dataset and five standard datasets of marathon video and natural scene text datasets show that the proposed method is better than the existing methods. However, it is noted from Fig.26 that the proposed method is not robust with respect to occlusion, severe blur and very small fonts while capturing images from long-distance cameras. Furthermore, if the human body does not expose skin and if text is present at the backside and/or on the legs of the human body, the proposed method does not perform well. Therefore, we plan to introduce new concepts such as context information based on human body structure without depending on skin for text and bib number detection in videos as our future work to handle such cases. Moreover, one can think of developing a new end-to-end deep learning architecture which can address the challenges of marathon, sports and natural scene images without segmenting torso regions like the proposed method.

## Acknowledgements

This work is supported by the National Natural Science Foundation of China under Grant No. 61672273, and No. 61832008, and also partially supported by Faculty Grant: GPF014D-2019, University of Malaya, Malaysia. We thank the authors of the paper [28] for sharing their dataset, which helped us to show that the proposed method is effective and useful. We also thank the anonymous reviewers for their constructive suggestions and comments to improve the quality and clarity of the work.